\documentclass[pdflatex,sn-mathphys-num]{sn-jnl}% Math and Physical 

\usepackage{graphicx}%
\usepackage{multirow}%
\usepackage{amsmath,amssymb,amsfonts}%
\usepackage{bm}%
\usepackage{amsthm}%
\usepackage{mathrsfs}%
\usepackage[title]{appendix}%
\usepackage{xcolor}%
\usepackage{textcomp}%
\usepackage{manyfoot}%
\usepackage{booktabs}%
\usepackage{algorithm}%
\usepackage{algorithmicx}%
\usepackage{algpseudocode}%
\usepackage{listings}%
\theoremstyle{thmstyleone}%
\theoremstyle{thmstyletwo}%

\theoremstyle{thmstylethree}%

\DeclareMathOperator*{\argmin}{arg\,min}
\usepackage{caption}
\begin{document}

\title[Article Title]{Counterfactual Explanation for Multivariate Time Series Forecasting with Exogenous Variables}


\author{\fnm{Keita} \sur{Kinjo}}\email{kkinjo@kyoritsu-wu.ac.jp}

\affil{\orgdiv{Faculty of Business}, \orgname{Kyoritsu Women's University}, \orgaddress{\street{2-2-1 Hitotsubashi}, \city{Chiyoda-ku}, \postcode{101-8437}, \state{Tokyo}, \country{Japan}}}


\abstract{Currently, machine learning is widely used across various domains, including time series data analysis. However, some machine learning models function as black boxes, making interpretability a critical concern. One approach to address this issue is counterfactual explanation (CE), which aims to provide insights into model predictions. This study focuses on the relatively underexplored problem of generating counterfactual explanations for time series forecasting. We propose a method for extracting CEs in time series forecasting using exogenous variables, which are frequently encountered in fields such as business and marketing. In addition, we present methods for analyzing the influence of each variable over an entire time series, generating CEs by altering only specific variables, and evaluating the quality of the resulting CEs. We validate the proposed method through theoretical analysis and empirical experiments, showcasing its accuracy and practical applicability. These contributions are expected to support real-world decision-making based on time series data analysis.}

\keywords{Counterfactual explanation, Interpretability, Time series forecasting, Exogenous variables}



\maketitle

\newpage 
\section{Introduction}\label{sec1}

Currently, artificial intelligence is being applied across various fields, with machine learning in particular seeing increasing adoption. Amid this trend, the accountability of artificial intelligence (AI) and machine learning systems has become a crucial issue. One major concern is the black-box nature of some machine learning models, which can make accurate predictions but fail to provide explanations for those predictions \citep{angelov2021explainable, minh2022explainable, dwivedi2023explainable, saeed2023explainable}. This lack of interpretability presents challenges not only for explaining predictions to humans but also for tuning models and making informed decisions based on them. Consequently, the fields of interpretable machine learning and explainable AI (XAI) have garnered significant attention, with various techniques and approaches being developed \citep{du2019techniques, molnar2020interpretable}. For example, some methods focus on using white-box models, whereas others aim to identify the key features that contribute to predictions.

There is a wide variety of temporally dependent data---known as time series data---such as sensor readings and economic indicators, such as stock prices. Both time series-specific methods and general machine learning approaches have been developed and applied to tasks such as forecasting. Numerous applications have demonstrated the effectiveness of applying machine learning models to such time series data. However, similar to other domains, the black-box nature of machine learning models has become a concern in time series analyses, highlighting the need for interpretability. To address this issue, various interpretability techniques have been proposed \citep{theissler2022explainable}. According to Theissler, methods have been developed at different temporal scales, including time point-based, subsequence-based, and instance-based approaches. For example, TimeSHAP extends the SHAP framework, which is commonly used in explainable AI, to identify the contribution of each feature and timestamp to a prediction based on time series data \citep{bento2021timeshap}. Other approaches aim to extract important sub-sequences or patterns that influence a model's prediction, thereby providing insights into the reasoning behind it \citep{spinnato2023understanding}.

Among these developments, counterfactual explanation (CE) is a promising approach for enhancing the interpretability of machine learning in time series analysis \citep{verma2024counterfactual, guidotti2024counterfactual}. A CE provides an answer to the question, “What minimal changes to the input would have altered the prediction?”—helping to identify influential variables and support decision-making. Traditionally, CE has been widely applied to domains, such as image and text data \citep{goyal2019counterfactual, ates2021counterfactual, prado2024survey, jeanneret2024text}. However, more recently, its application to time series data has gained increasing attention. In particular, there is a growing demand for CE methods that can be adapted to the unique characteristics of time series data, such as temporal dependencies and sensitivity to outliers. This study focused on this emerging area. 

Research on CEs in time series data can be broadly categorized into three areas based on differences in problem formulation.

The first area concerns CEs in time series classification. This involves determining how to modify a time series instance such that the predicted class changes to the desired target class. Such approaches are particularly useful in domains like electrocardiogram (ECG) classification. Numerous studies have investigated this topic. Early methods included \textit{CoMTE} \citep{ates2021counterfactual}, which generated realistic CEs based on existing data, and \textit{Native Guide} \citep{delaney2021instance}. Other approaches utilize local patterns or features in time series data, such as shapelets, motifs, and discords, and the rules among them \citep{li2022sgcf, bahri2022shapelet, refoyo2024subspace}. Some methods exploit global signal features, such as those extracted via the Discrete Fourier Transform, to generate invariant CEs \citep{bahri2025denoising}. \textit{TimeX} \citep{filali2022barycenter} uses Dynamic Time Warping (DTW) to compute the loss functions for CE generation. Additionally, optimization-based approaches, including evolutionary algorithms and multi-objective optimization, have been developed to improve CE quality \citep{hollig2022tsevo, huang2024txgen, refoyo2024multispace}. Techniques adapted from model interpretability, such as saliency maps and LIME, have also been applied to methods, such as \textit{CELS}, \textit{M-CELS} \citep{li2023cels, li2024mcels}, and \textit{Glacier} \citep{wang2024glacier}. Beyond these, model-agnostic approaches such as \textit{CFWoT} leverage reinforcement learning to dynamically generate CEs \citep{sun2024counterfactual}, whereas \textit{CounTS} \citep{yan2023selfinterpretable} utilizes Pearl's causal inference framework to estimate the effect of exogenous variables during CE generation. Collectively, these studies reflect the diversity and richness of the CE approaches for time series classification.

The second area concerns CEs for time series forecasting. This involves identifying how the input data should be modified such that the predicted values of a time series approach the desired target values. One of the few studies in this area is that of Wang et al. (2023), who formulate the problem as ensuring that the predicted values of a univariate time series fall within predefined upper and lower bounds \citep{wang2023forecasting}. They proposed an algorithm called \textit{ForecastCF} that solves this problem using gradient-based perturbations. However, their work also highlighted several open challenges, including extensions to multivariate time series and the incorporation of exogenous variables. 

The third area involves CEs in time series anomaly detection. This line of research is closely related to CEs for time series classification but focuses specifically on identifying the causes of anomalies. The core idea is to modify the original time series data slightly to generate samples that are no longer classified as anomalous, thereby revealing the factors responsible for this anomaly \citep{sulem2022diverse}. In addition to these categories, CE studies related to time series assume a Markov decision process (MDP) framework, focusing on counterfactual reasoning in sequential decision-making under uncertainty \citep{tsirtsis2021counterfactual}. 

To date, most existing studies on counterfactual explanations have focused on time series classification. In particular, substantial attention has been paid to extracting meaningful features from time series data and developing tailored methods for CE generation and model estimation. However, the application of CEs to time series forecasting remains limited and underexplored. As Wang et al. (2023) pointed out, there is a notable lack of research on counterfactual explanations in forecasting settings that involve multivariate time series or include exogenous variables. Moreover, few studies have considered optimizing exogenous variables across multiple time steps to guide the predicted time series toward a desired multistep target.

While nonlinear and nonparametric time series models based on machine learning often achieve high predictive accuracy, they differ from traditional white-box models in that the influence of exogenous variables can vary across time points, and the extent to which each variable affects the target variable is not explicitly known. In such settings, identifying which exogenous or intervention variables should be adjusted to bring the target variable closer to the desired value is critically important for both decision-making and model interpretability. 
For example, consider time series forecasting of sales in a marketing context. To achieve a target sales trajectory over multiple time steps, it is crucial to accurately assess the effects of exogenous variables such as advertising and determine their optimal values. This approach enables more practical and interpretable time series forecasting. 

To address this challenge, we propose a method for extracting \textit{Counterfactual Explanations in Time series forecasting with eXogenous variables (CET-X)}. In this approach, we assume that the value of the target variable is influenced by its own past values as well as exogenous variables and apply various predictive models under this assumption. We then generate counterfactual explanations by adjusting the exogenous variables such that the predicted values of the target variable across multiple time steps approach a specified target trajectory. Furthermore, we propose a method for capturing overall trends by aggregating the CEs extracted across the time series. In addition, we introduced evaluation metrics to analyze the properties and quality of the generated counterfactuals.
Building on the above, this study evaluated the effectiveness of the proposed method using both simulation data and real-world datasets. Through these experiments, we aim to assess the practical applicability of the method under realistic data conditions and demonstrate the significance of counterfactual explanations in time series forecasting.

The remainder of this paper is organized as follows. Section~\ref{sec2} introduces the proposed method, Section~\ref{sec3} presents experimental evaluations to validate its effectiveness, and Section~\ref{sec4} provides a discussion based on the results.

\section{Method}\label{sec2}

Section~\ref{sec2.1} describes our proposed method. Section~\ref{sec2.2} presents a technique for identifying important variables based on the proposed approach, and Section~\ref{sec2.3} introduces evaluation metrics for analyzing the characteristics of counterfactual explanations.

\subsection{Proposed method}\label{sec2.1}

This section introduces the data used in this study, general modeling framework, and formulation of the CE extraction problem. This series of methods is referred to as CET-X.

\subsubsection{Data and General Time Series Models}\label{sec2.1.1}

We consider a univariate time series for the target variable:
\[
X = \left\{ x_t \in \mathbb{R} \,\middle|\, t = 1, 2, \ldots, T \right\}.
\]
Similarly, the set of exogenous time series that influence \( X \) is defined as
\[
Z = \left\{ Z_k \,\middle|\, k = 1, 2, \ldots, K \right\},
\]
where each element \( Z_k \) represents one exogenous variable given by
\[
Z_k = \left\{ z_{k,t} \in \mathbb{R} \,\middle|\, t = 1, 2, \ldots, T \right\}.
\]
Here, \( k \) indexes the exogenous variables, and \( K \) denotes their total number.For simplicity, we refer to the target and exogenous variables as \( X \) and \( Z \), respectively.

We assume that variable \(x_t\) is influenced by its own past \(m\) time steps and by the past \(n\) time steps of exogenous variables \(Z_k\).  
The model was formulated as follows:
\begin{equation}
x_t = f\left(x_{t-1}, \ldots, x_{t-m}, z_{1,t-1}, \ldots, z_{1,t-n}, \ldots, z_{K,t-1}, \ldots, z_{K,t-n}\right) + \varepsilon_t,
\label{eq:model}
\end{equation}
where \( f:\mathbb{R}^{m+nK} \rightarrow \mathbb{R} \) represents a function estimated by long short-term memory (LSTM) or other machine learning methods.  
The noise term \(\varepsilon_t\) follows a normal distribution \(N(0, \sigma^2)\).  
In this study, we assumed no autocorrelation in the noise.

This formulation is general and can accommodate models such as auto-regressive model with exogenous inputs (ARX model), various time series neural network models, and standard machine learning algorithms.  
The lag lengths \(m\) and \(n\) were automatically selected based on prediction accuracy using a training/testing data split.

\subsubsection{Prediction and Counterfactual Explanation Extraction Problem}\label{sec2.1.2}

In this study, we formulated a counterfactual explanation extraction problem. 
To bring the predicted values \(\hat{x}_t\) from period \(T-q\) to \(T\) closer to the target values \(\bar{x}_t\), 
we manipulated the exogenous variables \(Z\) in the most recent \(q\) steps as the intervention variables.

Specifically, we aim to find the intervention variables:
\[
\widetilde{Z}_{T,q} = 
\left\{
\widetilde{z}_{1,T-1}, \ldots, \widetilde{z}_{1,T-q}, \ldots, 
\widetilde{z}_{K,T-1}, \ldots, \widetilde{z}_{K,T-q}
\right\}
\]
that minimize the following objective function:

\begin{equation}
\widetilde{Z}_{T,q}^{\ast}
=
\argmin_{\widetilde{Z}_{T,q}}
\left\{
L\left(\widetilde{Z}_{T,q}\right)
=
\left[
\sum_{t=T-q}^{T} 
w_t
\left(\bar{x}_t - \hat{x}_t\right)^{2}
\right]
+
\lambda\, d\left(\widetilde{Z}_{T,q}, Z_{T,q}\right)
\right\}
\label{eq:objective}
\end{equation}

\noindent
where \(\bar{x}_t\) are the desired target values and \(w_t (\geq 0)\) are time-dependent weights.  
Flexibility exists in the choice of weights. For instance, one can use exponentially decaying weights, 
assign all weights as one to compute a simple average, or set only the final time point weight as one 
in order to evaluate only \(\hat{x}_T\).  
The predicted value \(\hat{x}_t\) at time \(t\) was obtained using the machine learning model \(f\) introduced earlier.  
The parameter \(\lambda \geq 0\) is a penalty term (regularization parameter) that controls 
the magnitude of the intervention's impact.  

\[
Z_{T,q} =
\left\{
z_{1,T-1}, \ldots, z_{1,T-q}, \ldots, 
z_{K,T-1}, \ldots, z_{K,T-q}
\right\}
\]
represents the original values of the exogenous variables.  
The function \(d\) denotes a distance function, which is often referred to as proximity and serves to ensure 
the plausibility and feasibility of the counterfactual explanations, as required by the CE framework 
\citep{wachter2017counterfactual}.  

For instance, a weighted Euclidean distance,
\[
\sum_{k=1}^{K}\sum_{t=T-q}^{T-1}
\sqrt{\delta_{k,t}\left(\widetilde{z}_{k,t}-z_{k,t}\right)^{2}}
\]
or a weighted sum of squared differences as a pseudo-metric,
\[
\sum_{k=1}^{K}\sum_{t=T-q}^{T-1}
\delta_{k,t}\left(\widetilde{z}_{k,t}-z_{k,t}\right)^{2}
\]
can be used.  
Here, \(\delta_{k,t} > 0\), which allows the introduction of adjustment costs that vary across time points 
and depend on the specific index \(k\) of the exogenous variables.

Here, we describe a specific calculation for \(\hat{x}_t\). The prediction is made using the estimated function \(f\), post-intervention input sequence \(\widetilde{Z}_{T,q}\), and the following data:
\[
\left\{x_{T-q-1}, \ldots, x_{T-q-m}\right\}, \quad
\left\{z_{1,T-q-1}, \ldots, z_{1,T-q-n}, \ldots, z_{K,T-q-1}, \ldots, z_{K,T-q-n}\right\}.
\]
It is also possible to use the sequentially predicted values 
\(\left\{\hat{x}_{T-q-1}, \ldots, \hat{x}_{T-q-m}\right\}\)
instead of 
\(\left\{x_{T-q-1}, \ldots, x_{T-q-m}\right\}\).

First, the predicted value at time \(T-q\) after the intervention is given by:
\begin{equation}
\hat{x}_{T-q} = f\left(
x_{T-q-1}, \ldots, x_{T-q-m}, 
z_{1,T-q-1}, \ldots, z_{1,T-q-n}, \ldots, 
z_{K,T-q-1}, \ldots, z_{K,T-q-n}
\right).
\label{eq:pred_tq}
\end{equation}

The predicted value \(\hat{x}_{T-q}\) is then used as an input to predict the next time point \(\hat{x}_{T-q+1}\), 
and this process was repeated sequentially until the final prediction at time \(T\) was obtained.  
The final prediction is given by:
\begin{equation}
\hat{x}_{T} = f\left(
\hat{x}_{T-1}, \hat{x}_{T-2}, \ldots, \hat{x}_{T-q}, 
\widetilde{z}_{1,T-1}, \ldots, \widetilde{z}_{1,T-q}, \ldots, 
\widetilde{z}_{K,T-1}, \ldots, \widetilde{z}_{K,T-q}
\right).
\label{eq:pred_final}
\end{equation}

\begin{figure}[htbp]
    \centering
    \includegraphics[width=0.9\textwidth]{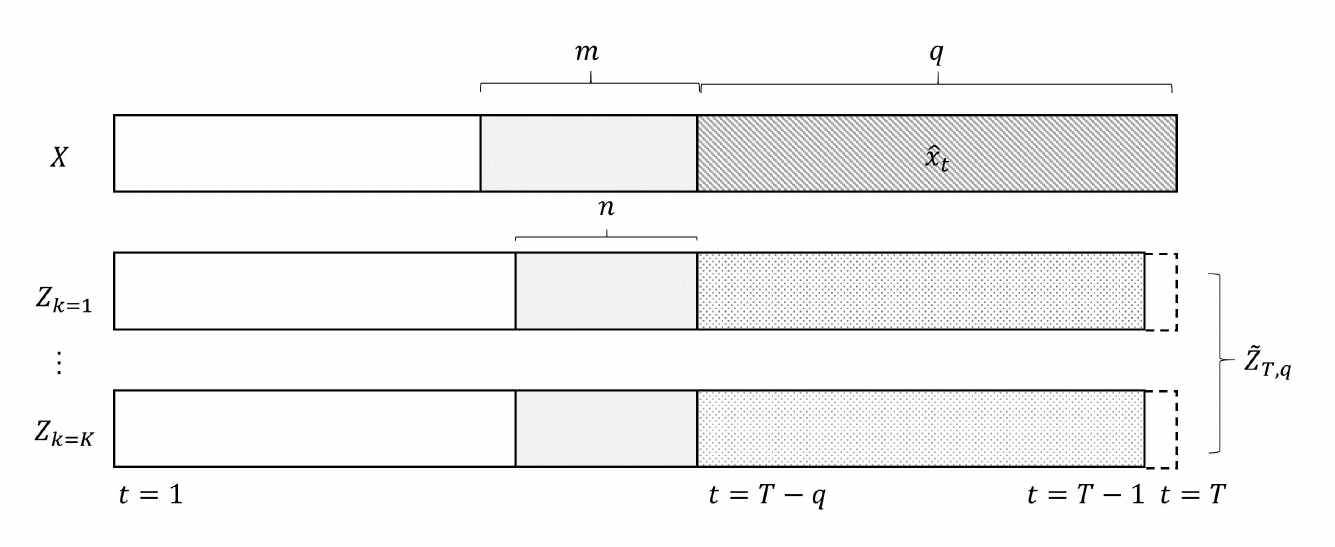}
    \caption{Overview of the proposed method.}
    \label{fig:overview}
\end{figure}

An overview of this process is shown in Figure~\ref{fig:overview}. 
Each box represents the time series data of \(X\) and \(Z_k\), with the horizontal axis indicating time \(t\).  
In this figure, the predicted value \(\hat{x}_{T-q}\) (indicated by shaded areas) is estimated based on the data of \(X\) from \(t = T-q-m\) to \(t = T-q-1\), and the data of \(Z_k\) from \(t = T-q-n\) to \(t = T-q-1\).  
Next, at time \(t = T-q\), the prediction \(\hat{x}_{T-q+1}\) (shaded area) is calculated using the intervened exogenous variables \(\widetilde{Z}_{T,q}\) (dotted background) and the original \(Z_k\) data from \(t = T-q-1\) to \(t = T-q-m+1\).  
This procedure is repeated sequentially to ultimately predict \(\hat{x}_T\) (shaded area).  
Based on this process, the goal is to determine \(\widetilde{Z}_{T,q}\) optimally such that the predicted values \(\hat{x}_t\) approach the target values. 
Optimization can be performed using methods such as stochastic gradient descent.

Here, we discuss the settings of the proposed method. 
This study limits the optimization range of \(Z\) to \(q\) time points. 
However, if the prediction of \(x_{t-q}\) is influenced by earlier values of \(Z\), 
it is theoretically possible to optimize the entire history of \(Z\) 
(by setting \(q = T - 1 - \max(m, n)\)). 
Nevertheless, there are several reasons for restricting the optimization to \(q\) time points. 
First, as the number of prediction steps increases, the accumulation of prediction errors may become significant. 
Second, if the entire history of \(Z\) is included, the dimensionality of the optimization variables increases substantially, which presents computational challenges. 
Finally, in real-world policy-making and marketing applications, it is often more practical to adjust only ``recent interventions.'' 
For instance, the result of changing all interventions over the past ten years is generally less interpretable than the result of adjusting interventions from just the past three months. 

However, it is also possible to target \(X\) after time \(T-q\) and obtain a solution that includes \(Z\) up to \(T-q\). 
In other words, the range of the target and the counterfactual instance can be shifted. 
In this case, by setting a larger value of \(q\) and modifying the corresponding \(w_t\) 
(e.g., setting it to zero), the optimization can still be performed within the same framework. 
Thus, by appropriately adjusting the configuration of \(w_t\), the range of optimization can be flexibly controlled.

Some exogenous variables in \(Z\), such as the temperature, cannot be controlled. 
Additionally, in practical applications such as marketing or policy-making, 
it is often desirable to focus solely on the effects of controllable interventions. 
In such cases, it is straightforward to restrict optimization to specific variables of interest. 
The details of this procedure are described in Section~\ref{sec2.2}.

In this study, unlike the approach taken by Wang et al.~\citep{wang2023forecasting}, 
we targeted past time points rather than future predictions, such as \(T+q\). 
This focus was primarily aimed at investigating past causes. 
Additionally, in future periods, the values of \(Z\) will not be available, making it impossible to calculate the distance to the counterfactual. 
However, our method can be readily extended to predictions by introducing several reasonable assumptions. 
This extension is discussed in detail in the Discussion section (Section~\ref{sec4}).

Here, we discuss the uniqueness of solution \(\widetilde{Z}_{T,q}\) in the problem. 
For the optimal solution to be uniquely determined, continuity and strong convexity (at least locally) of the objective function \(L\left(\widetilde{Z}_{T,q}\right)\) are the required conditions. 
The objective function consists of the sum of the prediction error, calculated using the target values and function \(f\), and the distance function \(d\). 
Therefore, it is necessary that \(f\) must be continuous and preferably locally strongly convex. 
This ensures that the error term provides sufficient curvature to contribute to the convexity of the overall objective function, which helps to prevent the occurrence of multiple solutions that yield the same minimum. 
For example, this condition is satisfied when \(f\) is a linear function and \(w_t \geq 0\). 
In addition, when using the weighted sum of squares as the distance function \(d\), as described earlier, the Hessian matrix of \(d\) is proportional to the identity matrix, which guarantees strong convexity. 
Specific cases where the optimal solution can be analytically computed are discussed in Section~\ref{sec3.1} and the Appendix\ref{secA1}.

Furthermore, if the parameter \(\lambda\) is set sufficiently large, even if the error term does not possess complete strong convexity, the positive curvature provided by the distance function \(d\) can significantly contribute to the Hessian matrix of the overall objective function, making the entire function strongly convex. 
Conversely, if the error term is sufficiently strongly convex, it is possible to set \(\lambda\) to a small value, allowing the curvature of the error term to dominate the overall objective function. 
In such cases, a unique minimum solution may still be obtained, even if \(d\) is nonconvex. 
However, when using complex models, such as machine learning, it is generally difficult to guarantee the uniqueness of the solution, because many of these conditions depend heavily on the properties of the function \(f\). 
Therefore, approximate optimization methods such as stochastic optimization are required to obtain practical solutions.

\subsection{Method for Extracting Important Features}\label{sec2.2}

The optimal solution \(\widetilde{Z}_{T,q}^{\ast}\) depends on the data level and patterns around time \(T\). 
Therefore, it is important to consider methods for analyzing the average influence of \(Z\) across the entire dataset. 
Such analyses can be useful for identifying invariant structures within the data and selecting more robust interventions in practical applications. 

As an overview of the proposed method, we slide the procedure described in Section~\ref{sec2.1}~(2) backward by one time step at a time and repeatedly apply it to extract counterfactual explanations across the entire time series. 
Specifically, the procedure consists of the following steps:

\renewcommand{\theenumi}{\roman{enumi}}
\begin{enumerate}
    \item Assume \(T-1\) is the final time point \(T\). At this point, the lower bound of the optimization window \(T-q\) and the data range \(T-q-\max{\left(m,n\right)}\) are reduced by one time step to \(T-q-1\) and \(T-q-\max{\left(m,n\right)}-1\), respectively. Subsequently, extract \(\widetilde{Z}_{T-1,q}^\ast\).

    \item Next, shift the final time point backward by one step again, assuming \(T-2\) as the new final time point, and repeat the same procedure to extract \(\widetilde{Z}_{T-2,q}^\ast\).

    \item Repeat this process \(j\) times until the condition \(T-q-\max{\left(m,n\right)}-j=1\) is satisfied and continue extracting \(\widetilde{Z}_{T-j,q}^\ast\).

    \item Calculate statistical measures such as the mean and standard deviation of the differences:
    \[
    \left\{
    \widetilde{Z}_{T-1,q}^\ast - Z_{T-1,q},\;
    \ldots,\;
    \widetilde{Z}_{T-j,q}^\ast - Z_{T-j,q}
    \right\}.
    \]
    When the dataset is large, it is not necessary to apply this procedure to all time intervals, and sampling can be used as an alternative.
\end{enumerate}

\subsection{Method for Extracting Counterfactual Explanations by Modifying Only Selected Exogenous Variables}
\label{sec2.3}

In real-world problems, it is often the case that not all exogenous variables \(Z_k\) can be controlled. 
In certain situations, it is desirable to modify only specific variables. 
For example, variables such as advertising can typically be controlled in marketing applications, whereas variables such as weather cannot.

Moreover, in models composed of multiple exogenous variables \(Z_k\), interactions between different \(Z_k\) variables as well as interactions with the target variable \(X\) may exist. 
To investigate these interaction effects, it is possible to compare CEs generated by modifying individual \(Z_k\) variables with those generated by modifying multiple \(Z_k\) variables simultaneously.

To this end, the optimization can be restricted to specific \(Z_k\) variables, and the corresponding counterfactuals \(\widetilde{Z}_{T,q,k}^\ast\) can be extracted accordingly:
\begin{equation}
\widetilde{Z}_{T,q,k}^{\ast}
=
\argmin_{\widetilde{Z}_{T,q,k}}
\left\{
L\left(\widetilde{Z}_{T,q,k}\right)
=
\left[
\sum_{t=T-q}^{T} w_t \left(\bar{x}_t - \hat{x}_t\right)^2
\right]
+
\lambda \, d\left(\widetilde{Z}_{T,q,k}, Z_{T,q,k}\right)
\right\}.
\label{eq:partial_opt2}
\end{equation}

\subsection{Evaluation Metrics}\label{sec2.4}

Finally, we describe the evaluation metrics of counterfactual explanations (CEs). 
Numerous studies have proposed various evaluation criteria for CEs \citep{verma2024counterfactual, guidotti2024counterfactual}. 
Metrics specifically designed for time series counterfactual explanations have also been proposed \citep{wang2023forecasting}. 
Based on these studies, we adopted the following evaluation metrics.  

First, we consider \textit{validity}, which measures how closely the predicted values of the target variable approach the desired values. 
In this study, because the weighting factor \(w_t\) was incorporated, we used the following definition based on the optimization formula. 
Generally, smaller validity values were preferred.

\begin{equation}
X\text{-loss} = \sum_{t=T-q}^{T} w_t \left( \bar{x}_t - \hat{x}_t \right)^2
\label{eq:Xloss}
\end{equation}

Next, to measure the \textit{proximity} between the counterfactual and the original data, 
we directly use the following \(Z\)-loss function. 
In general, smaller proximity values are preferred.

\begin{equation}
Z\text{-loss} = \sum_{k=1}^{K} \sum_{t=T-q}^{T-1} 
\left( \widetilde{z}_{k,t} - z_{k,t} \right)^2
\label{eq:Zloss}
\end{equation}

Furthermore, we use the \textit{Total loss}, which incorporates the balance between these components 
through the parameter \(\lambda'\). 
A lower total loss is also preferred.

\begin{equation}
\text{Total loss} = X\text{-loss} + \lambda' Z\text{-loss}
\label{eq:Totalloss}
\end{equation}

In addition, when CEs involve controllable variables, their values are often interdependent and may not change drastically over time. 
For instance, in marketing, sudden adjustments to prices or budgets are typically difficult. 
From this perspective, we also propose \textit{temporal smoothness} (TS), 
which measures the time-wise smoothness of each exogenous variable \(k\) in the CE. 
A lower TS value indicates a smoother, more realistic transition over time.

\begin{equation}
\text{TS} = 
\sum_{k=1}^{K} \sum_{t=T-q}^{T-3} 
\left| 
\left( \widetilde{z}_{k,t+2} - \widetilde{z}_{k,t+1} \right) 
- 
\left( \widetilde{z}_{k,t+1} - \widetilde{z}_{k,t} \right)
\right|
\label{eq:TS}
\end{equation}

Finally, we proposed an evaluation metric to measure the difference between the theoretical and numerically computed solutions of \(\widetilde{Z}_{T,q}\), as discussed in Section~\ref{sec2.1}. 
This metric is applicable only in specific situations in which the solution is unique, and the theoretical values \(z_{k,t}^{\text{true}}\) can be explicitly calculated. 
The mean absolute error (MAE) was used to quantify the difference between the theoretical and numerical solutions of the CE. 
A lower MAE indicates a higher accuracy.

\begin{equation}
\text{MAE} = 
\frac{
\sum_{k=1}^{K} \sum_{t=T-q}^{T-1} 
\left| z_{k,t}^{\text{true}} - \widetilde{z}_{k,t} \right|
}{
K + q
}
\label{eq:MAE}
\end{equation}

\section{Experiment}\label{sec3}

Subsequently, we conducted experiments using the proposed CET-X method. 
In Section~\ref{sec3.1}, we use simulation data to evaluate how the CEs generated by the proposed method vary depending on different values of \(\lambda\), \(q\), and the choice of optimization algorithm. 
We also assessed the accuracy of the method by comparing the numerical solutions with the analytically derived \(Z\) and by examining whether the method could correctly estimate the predefined causal effects. 
In Section~\ref{sec3.2}, the practical usefulness of the proposed method is validated using real-world datasets.

\subsection{Simulation Data}\label{sec3.1}
\subsubsection*{(1) Data Based on the ARX Model}\label{sec3.1.1}

We now describe the example used in the simulation. 
As a simple case, we consider a model where the target variable is influenced by its own one-period lag \(x_{t-1}\) 
and two exogenous variables \(z_{t-1,k}\).

\begin{equation}
x_t = \alpha x_{t-1} + \sum_{k=1}^{2} \beta_k z_{t-1,k} + \varepsilon_t
\label{eq:arx}
\end{equation}

The target variable \(x_t\) is a time series with an initial value of 0. 
The exogenous variables \(z_{t-1,k}\) consist of two independent variables, 
each following a standard normal distribution \(N(0, 1^2)\). 
The parameter \(\alpha\) is set to 0.6, and the weights \(\beta_k\) are 
\(\beta_1 = 0.2\) and \(\beta_2 = 0.5\). 
The noise term \(\varepsilon_t\) follows a normal distribution \(N(0, 0.1^2)\). 
In total, 200 data points were generated using this model. 
The generated data have a mean of \(-0.002\) and a variance of \(0.654\).

This is a linear model, and the analytical solution for the optimal counterfactual \(\widetilde{Z}_{T,q}^{\ast}\) can be computed. 
By sequentially arranging the optimal values from past to present, the solution can be expressed as a vertical vector \(\operatorname{vec}(\widetilde{Z}_{T,q}^{\ast})\).

\begin{equation}
\operatorname{vec}(\widetilde{Z}_{T,q}^{\ast})
=
\left( B_{\mathrm{opt}}^{T} W B_{\mathrm{opt}} + \lambda I \right)^{-1}
\left(
B_{\mathrm{opt}}^{T} W
\left( \bar{\boldsymbol{x}} - A \boldsymbol{x} - B_{\mathrm{fix}} Z_{\mathrm{fix}} \right)
+ \lambda Z_{\mathrm{opt}}
\right)
\label{eq:analytic}
\end{equation}

\(\boldsymbol{x}\) is a \((q+1)\times1\) matrix consisting only of the initial value \(x_{t-q-1}\), 
\(\bar{\boldsymbol{x}}\) is a \((q+1)\times1\) matrix composed only of the target values. 
\(W\) is a \((q+1)\times(q+1)\) matrix where the weights \(w_t\) are arranged diagonally in order from the past. 
\(A\) is a \((q+1)\times(q+1)\) matrix composed of \(\alpha\). 
\(Z_{\mathrm{fix}}\) is a \(2\times1\) matrix of the exogenous variables \(Z_{t,k}\) that are not subject to optimization. 
\(Z_{\mathrm{opt}}\) is a \((2q)\times1\) matrix of the exogenous variables \(Z_{t,k}\) that are subject to optimization. 
\(B_{\mathrm{fix}}\) is a \((q+1)\times2\) coefficient matrix composed of \(\alpha\) and \(\beta_k\) for the non-optimized \(Z_{\mathrm{fix}}\). 
\(B_{\mathrm{opt}}\) is a \((q+1)\times(2q)\) coefficient matrix composed of \(\alpha\) and \(\beta_k\) for the optimized \(Z_{\mathrm{opt}}\). 
\(I\) is the identity matrix. 
Please refer to the Appendix\ref{secA1} for further details. 

The models used in this study are as follows: 
ARX model, 
multi-layer perceptron (MLP), recurrent neural network (RNN), 
long short-term memory (LSTM) \citep{hochreiter1997long} with a hidden dimension of 8, 
and gated recurrent units (GRU) \citep{chung2014empirical} with a hidden dimension of 8. 
For model training, we employed a stochastic gradient descent (SGD), which is commonly used in machine learning.

Specifically, we construct a dataset consisting of pairs of target variables \(x_t\) and explanatory variables 
\(\left\{ x_{t-1}, \ldots, x_{t-m}, z_{1,t-1}, \ldots, z_{1,t-n}, \ldots, z_{K,t-1}, \ldots, z_{K,t-n} \right\}\). 
This dataset was divided into training and test sets. 
Model training was performed on the training set, and prediction accuracy was evaluated on the test set. 
The proportion of the training data was set to 80\% of the total dataset, corresponding to the first 80\% of the time series from the starting point.

First, we compared the models and selected the best-performing one (Table~\ref{tab:model_comparison}). 
We explored combinations of \(m\) and \(n\) where \(m = 1, 2, 3\) and \(n = 1, 2, 3\), and presented the top five results. 
The results clearly show that the original model, which assumes a lag of 1 for \(X\) and a lag of 1 for \(Z\), was successfully identified.

\begin{table*}[htbp]
\centering
\caption{Model Comparison (Linear): Top 5 Results.}
\label{tab:model_comparison}
\vspace{2mm}
\includegraphics[width=0.5\textwidth,keepaspectratio]{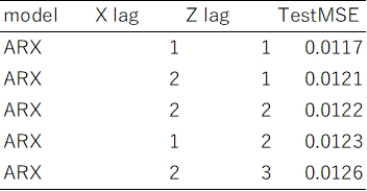}
\end{table*}

The target value \(\bar{x}_t\) was set to 2, which was larger than the mean of the data. 
We also prepare three types of weights \(w\) as follows:

\[
w_0 = \left( \frac{1}{q+1}, \frac{1}{q+1}, \ldots, \frac{1}{q+1} \right),
\]

\[
w_1 = \left( \frac{r}{sr}, \frac{r^2}{sr}, \ldots, \frac{r^{q+1}}{sr} \right),
\quad \text{where } sr = r + \ldots + r^{q+1}, \text{ and } r = 0.5,
\]

\[
w_2 = \left( 0, 0, \ldots, 0, 1 \right).
\]

All of these are vectors of length \(q+1\).

Next, we compare various evaluation metrics across different values of 
\(\lambda \in \left\{ 0.1,\, 0.5,\, 1.0,\, 2.0,\, 3.0,\, 5.0 \right\}\) 
with \(q = 3\) fixed, 
and across different values of 
\(q \in \left\{ 3, 4, 5, 6, 7 \right\}\) 
with \(\lambda = 3\) fixed. 

\begin{figure}[htbp]
\centering
\includegraphics[width=0.85\linewidth,keepaspectratio]{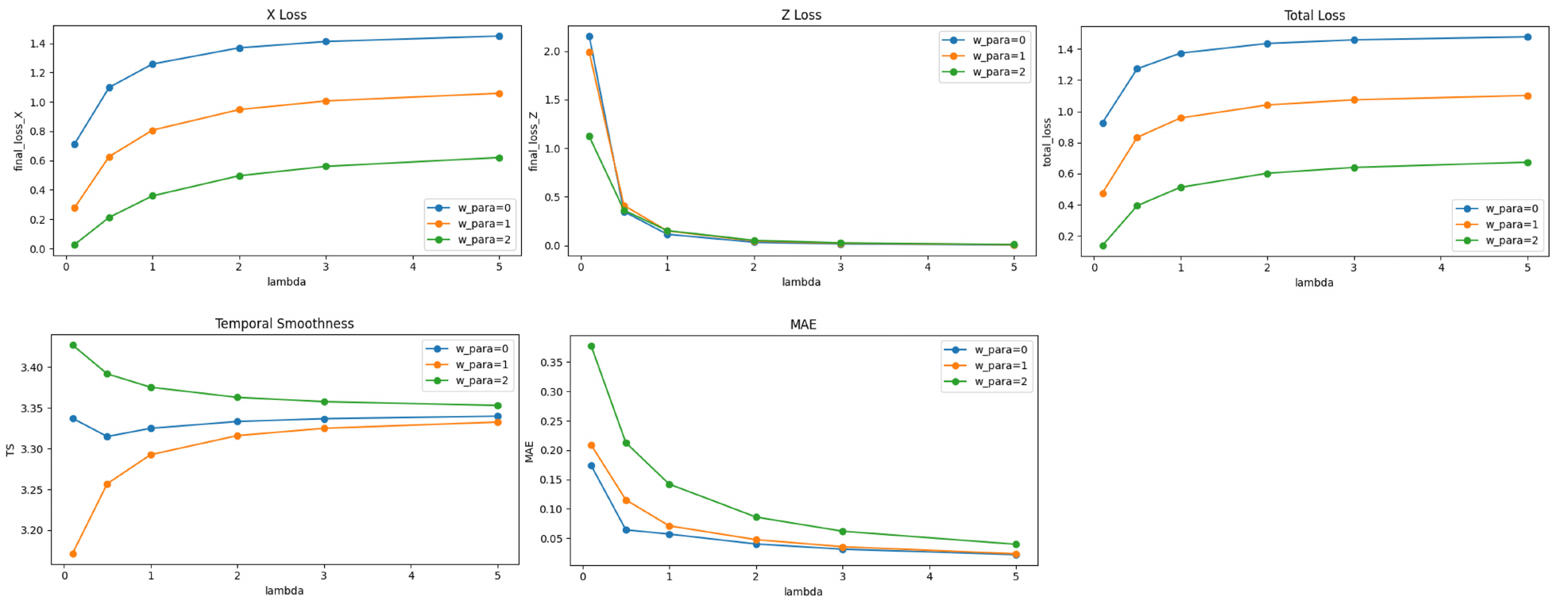}
\caption{Comparison of Different \(\lambda\) (Linear Case).}
\label{fig:lambda_comparison}
\end{figure}

We examined the effect of \(\lambda\) (Figure~\ref{fig:lambda_comparison}). 
In all weight settings \(w\), it is evident that as \(\lambda\) increased, 
the loss for \(X\) increased while the loss for \(Z\) decreased. 
This trend was particularly prominent when large adjustments were made, as in the case of \(w_0\). 

Additionally, the TS decreased as \(\lambda\) increased under \(w_2\), 
indicating a transition from a simple solution to a smoother one. 
Conversely, under \(w_0\), TS increased as \(\lambda\) increased, 
suggesting a transition from a smooth solution to a more complex one.

The MAE, which represents the difference from the theoretical values, was generally low, indicating that the estimates were appropriate. 
Furthermore, MAE tends to decrease when \(\lambda\) exceeds a certain threshold. 
However, for \(w_2\), the estimation error was relatively large. 
This is likely because \(w_2\) only targets the final point of \(X\), which can cause deviations in the intermediate values of \(X\) and \(Z\).

\begin{figure}[htbp]
\centering
\includegraphics[width=0.85\linewidth,keepaspectratio]{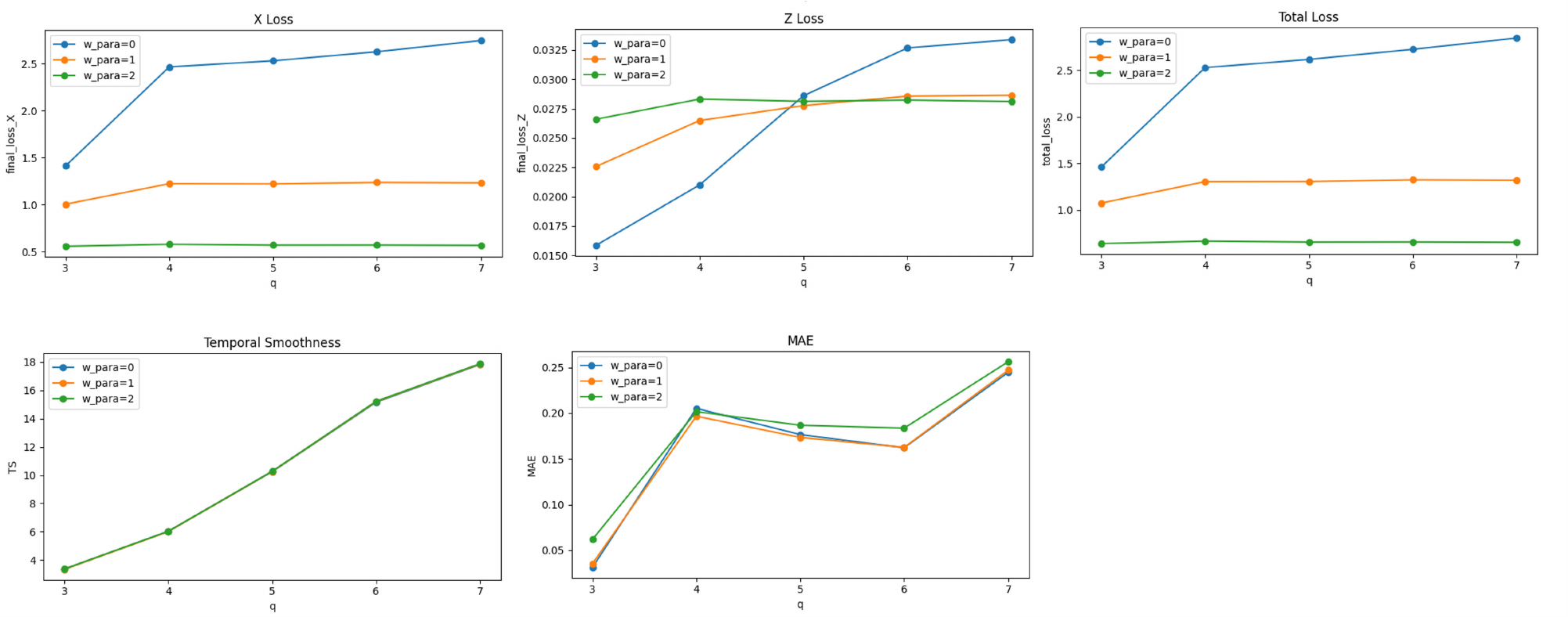}
\caption{Comparison of Different \(q\) (Linear Case).}
\label{fig:q_comparison}
\end{figure}

Next, we examined the effect of \(q\) (Figure~\ref{fig:q_comparison}). 
All losses—the loss for \(X\), the loss for \(Z\), and the Total loss—increase as \(q\) increases. 
However, the variation is relatively small for \(w_2\), suggesting that the optimal \(Z\) adjusted for a larger \(q\) does not significantly alter the predicted values of \(X\). TS consistently worsens with increasing \(q\). The MAE remains generally low, indicating an accurate overall estimation. 
However, it exhibited a pattern of first worsening, then improving, and worsening again. 
This suggests that there may be an optimal value of \(q\) depending on the specific problem setting. 
One possible reason is that the relative effect of the penalty parameter \(\lambda\) may vary greatly depending on the value of \(q\).

Thus, the combination of \(\lambda\), \(q\), and \(w\) has a significant impact on each evaluation metric. 
Therefore, it is important to select an appropriate balance of \(\{\lambda,\, q,\, w\}\) depending on the problem setting and operational policy.

We compared the optimization methods. 
The learning rate for SGD is set to 0.01, and the differences with respect to the momentum parameter 
\( \text{momentum} \in \{0.1,\,0.5,\,0.9\} \) 
are summarized in Table~\ref{tab:optimization_methods}. 
The momentum is a parameter that helps smooth the learning process by incorporating past gradient information. 
It carries over the previous direction of movement, making the trajectory smoother and potentially accelerating the convergence to the optimal solution. 
When the momentum was set to 0, the algorithm became standard SGD. 
If the value is small, the ability to escape from local optima tends to weaken.

As shown in the table, when the momentum is low, the loss function for \(Z\) decreases, whereas when the momentum is high, the loss function for \(X\) decreases. 
Furthermore, in more challenging settings where multiple \(X\) values (such as in \(w_0\) and \(w_1\)) are guided toward the target value, the total loss tends to be smaller when the momentum is higher. 
Thus, not only does the use of SGD matter but fine-tuning its parameters can also lead to more optimal outcomes.

\begin{table*}[htbp]
\centering
\caption{Comparison of Optimization Methods (Linear Case).}
\label{tab:optimization_methods}
\vspace{2mm}
\includegraphics[width=0.9\textwidth,keepaspectratio]{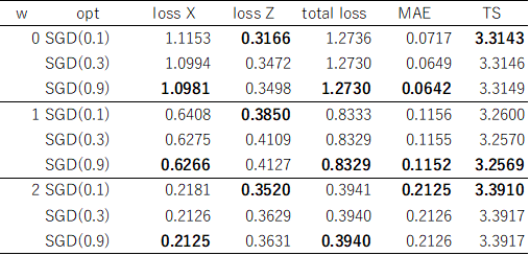}
\end{table*}

Finally, we describe the extraction of important features over the entire time series, as discussed in Section~2.2 (Table~\ref{tab:important_features}). 
As shown below, the mean change in \( z_{t-1,k=2} \) was larger than that in \( z_{t-1,k=1} \), and this tendency was consistently observed across all time points. 
The fact that the impact of \( k=2 \) is stronger and more positive is consistent with the characteristics of the original simulated data, 
indicating that the proposed method can successfully identify important features.

\begin{table*}[htbp]
\centering
\caption{Comparison of Important Features over the Entire Time Series (Linear Case).}
\label{tab:important_features}
\vspace{2mm}
\includegraphics[width=0.9\textwidth,keepaspectratio]{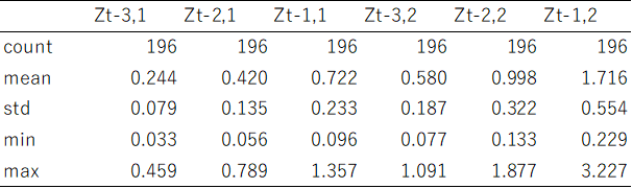}
\end{table*}

\subsubsection*{(2) Data Incorporating Nonlinearity and Interactions}\label{sec3.1.2}

Next, we consider a model that incorporates nonlinearities, such as \(\tanh\), power functions, and interaction terms. 
The analytical solution of this model is unknown.

\begin{equation}
x_t = 0.3\tanh{\left(x_{t-1}\right)} 
+ 0.1x_{t-2}^{1.5} 
+ \sum_{k=1}^{2}\left(\beta_k z_{t-1,k}^2\right)
+ 0.05\sum_{k=1}^{2}\left(x_{t-1}z_{t-2,k}\right)
+ \varepsilon_t
\label{eq:nonlinear_model}
\end{equation}

Variable \(x_t\) represents the time series data, and all initial values are set to zero. 
The exogenous variables \(z_{t-1,k}\) and \(z_{t-2,k}\) are defined in the same way as before, consisting of two types and following a standard normal distribution \(N(0,1^2)\). 
The weights \(\beta_k\) are given as \(\beta_1=0.2\) and \(\beta_2=0.5\). 
The noise term \(\varepsilon_t\) follows a normal distribution \(N(0,0.1^2)\). 
A total of 200 data points were generated with a mean of 0.963 and a variance of 0.750.  
The models were compared (Table~\ref{tab:model_comparison_nonlinear}). 
The results showed that the LSTM model with a lag of 2 for \(X\) and 1 for \(Z\) performed the best.

\begin{table*}[htbp]
\centering
\caption{Model Comparison (Nonlinear): Top 5 Results.}
\label{tab:model_comparison_nonlinear}
\vspace{2mm}
\includegraphics[width=0.5\textwidth,keepaspectratio]{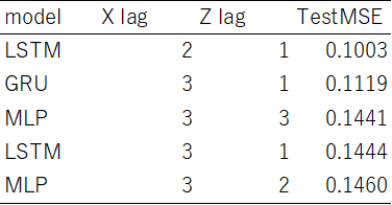}
\end{table*}

Next, we compare various evaluation metrics across different values of 
\(\lambda \in \{0.1, 0.5, 1.0, 2.0, 3.0, 5.0\}\) 
with \(q=3\) fixed, and across different values of 
\(q \in \{3, 4, 5, 6, 7\}\) 
with \(\lambda=3\) fixed.

Regarding \(\lambda\), it is evident that in all weight settings \(w\), 
as \(\lambda\) increased, the loss for \(X\) increased, while the loss for \(Z\) decreased.
This trend was more pronounced in cases in which large changes were allowed, such as with \(w_0\).  
Additionally, for \(w_2\) the TS decreased as \(\lambda\) increased, 
indicating a shift from a simple solution to a smoother one. 
In contrast, for \(w_0\), TS increased with increasing \(\lambda\), 
suggesting a shift from a smooth solution to a more complex one.

Next, looking at \(q\), both the \(X\) loss and the Total loss increased as \(q\) increased, 
whereas the \(Z\) loss decreased. 
This suggests that, in nonlinear models, increasing \(q\) allows the use of earlier values, 
which may have a greater influence on \(X\) with smaller changes in \(Z\). 
In nonlinear structures, the selection of an appropriate value for \(q\) is important. 
In addition, the TS consistently worsened as \(q\) increased.

\begin{figure*}[htbp]
    \centering
    \includegraphics[width=0.9\textwidth, keepaspectratio]{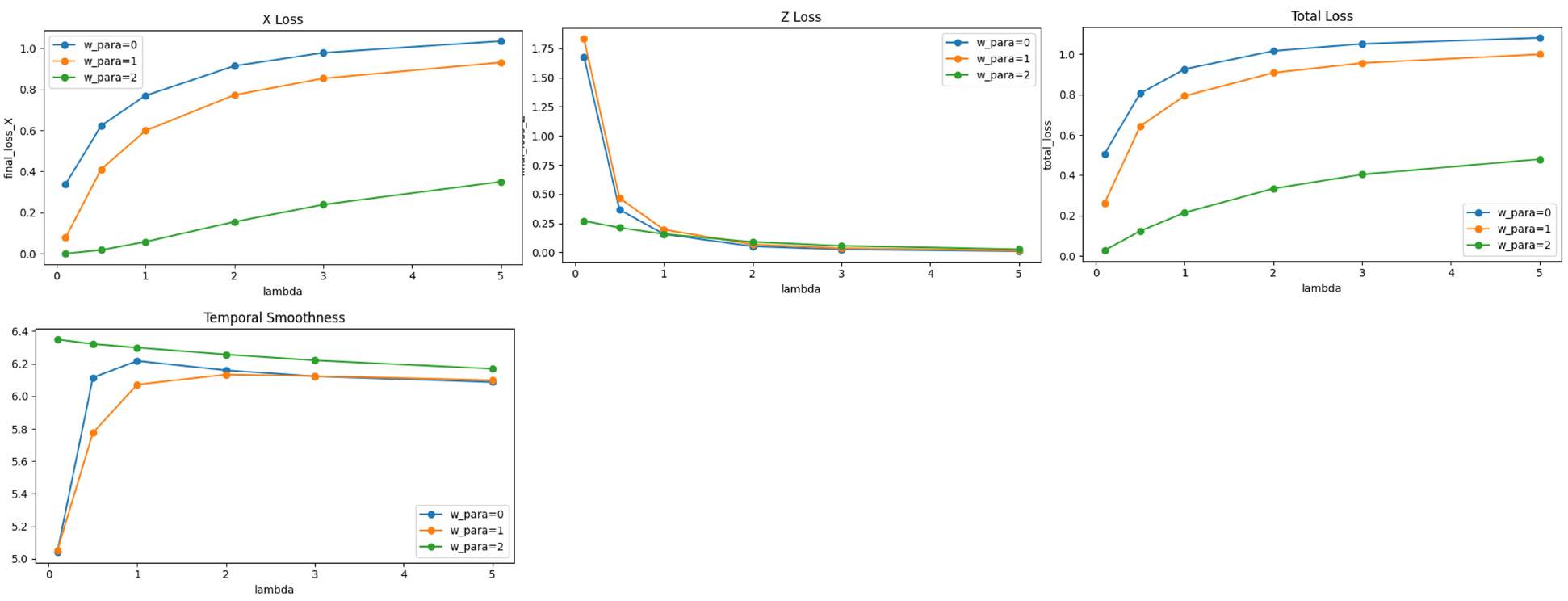}
    \caption{Comparison of different \(\lambda\) values (nonlinear case).}
    \label{fig:lambda_nonlinear}
\end{figure*}

\begin{figure*}[htbp]
    \centering
    \includegraphics[width=0.9\textwidth, keepaspectratio]{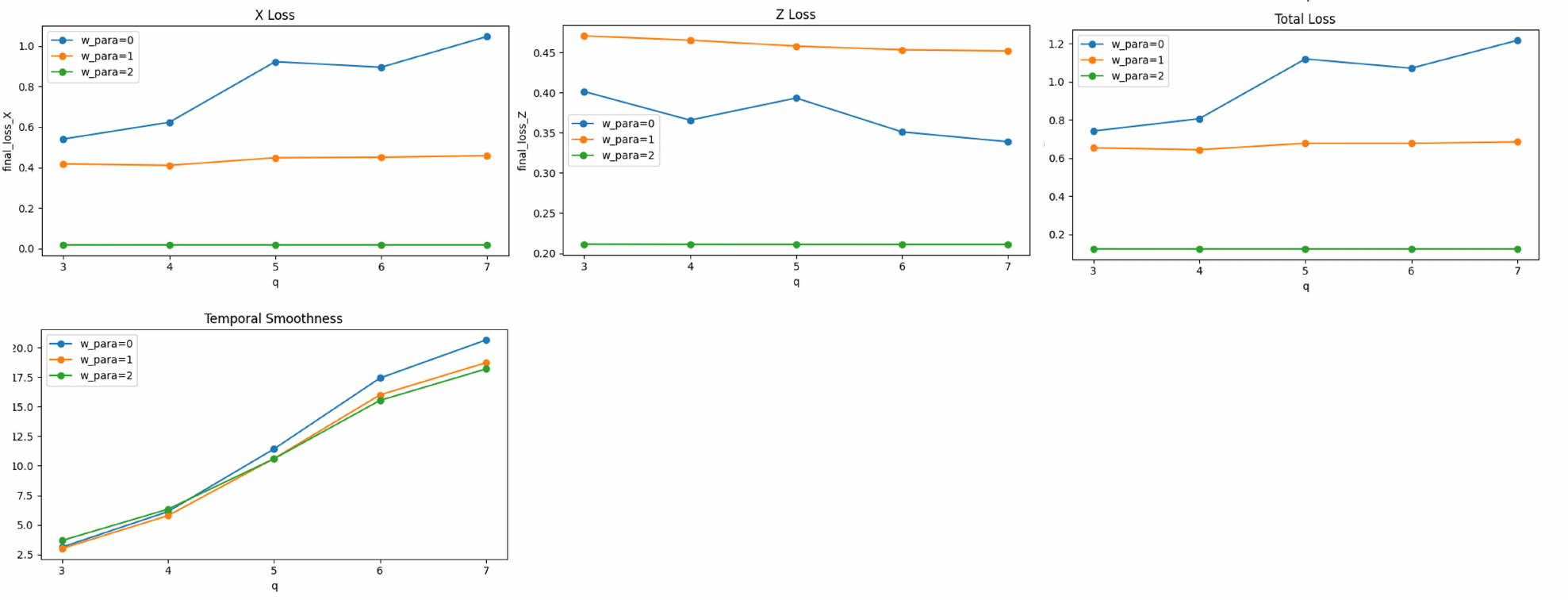}
    \caption{Comparison of different \(q\) values (nonlinear case).}
    \label{fig:q_nonlinear}
\end{figure*}

We compared the optimization settings. 
As shown in Table~\ref{tab:optimization_nonlinear}, 
when the momentum was small, the loss for \(Z\) is reduced, 
whereas when the momentum was large, the loss of \(X\) decreased. 
Furthermore, in more challenging scenarios, such as when using \(w_0\) or \(w_1\), 
where multiple values of \(X\) are guided toward the target, 
a larger momentum results in a smaller total loss.

\begin{table*}[htbp]
\centering
\caption{Comparison of Optimization Methods (Nonlinear Case).}
\label{tab:optimization_nonlinear}
\vspace{2mm}
\includegraphics[width=0.8\textwidth,keepaspectratio]{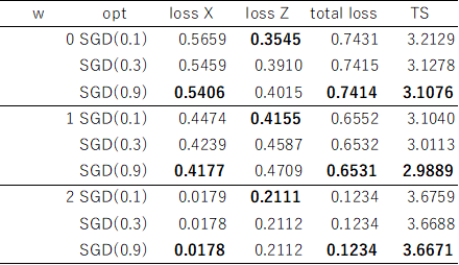}
\end{table*}

Next, we examined the important variables (Table~\ref{tab:important_features_nonlinear}). 
As shown below, the average values were generally close to zero, 
with the minimum values being negative and the maximum values being positive. 
This indicates that there were no features with consistently positive or negative effects. 
Moreover, the standard deviations for \(z_{t-3,k=2}\), \(z_{t-2,k=2}\), and \(z_{t-1,k=2}\) are relatively high. 
This suggests that the influence of these variables varied significantly depending on the data, 
implying strong effects. 
This is consistent with the observation that \(\beta_2=0.5\).

\begin{table*}[htbp]
\centering
\caption{Comparison of Important Features over the Entire Time Series (Nonlinear Case).}
\label{tab:important_features_nonlinear}
\vspace{2mm}
\includegraphics[width=0.9\textwidth,keepaspectratio]{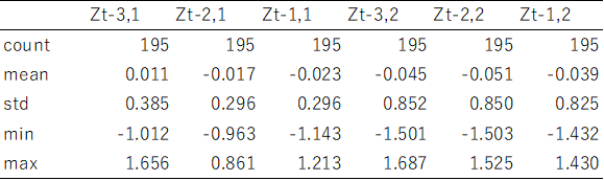}
\end{table*}

\subsection{Real-World Data}\label{sec3.2}

Next, we evaluated the effectiveness of CET-X using real-world data. 
In this study, we analyzed the search trends for specific language terms using Google Trends. 
Specifically, we use the monthly search volume for ``sake'' (Japanese rice wine) as the target variable, 
and the search volumes for ``washoku'' (Japanese cuisine, \(k=1\)) and ``nabe'' (Japanese hot pot, \(k=2\)) as exogenous variables. 
Data were collected in Japanese from February 2020 to February 2025 for a total of 61 time points. 
The original data were normalized relative values that varied by region and time for each variable.

\begin{figure*}[htbp]
    \centering
    \includegraphics[width=0.8\textwidth, keepaspectratio]{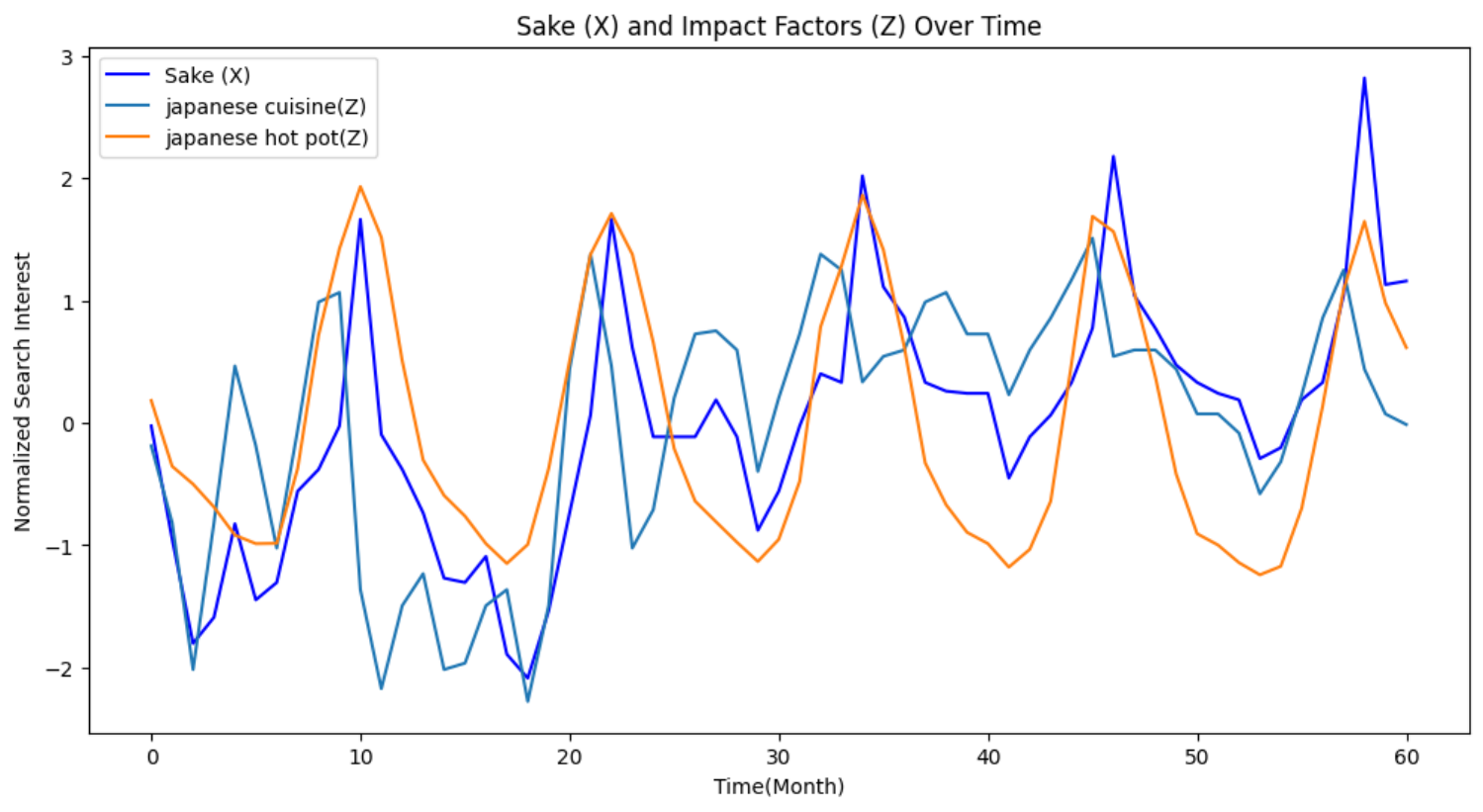}
    \caption{Standardized original data.}
    \label{fig:standardized_data}
\end{figure*}

First, the descriptive statistics of the time series data are shown in Table~\ref{tab:descriptive_real}. 
Notably, the average search volume and standard deviation for \textit{Japanese hot pot} are relatively high. 
Next, the standardized values of these time series are plotted in the figure below (Figure~\ref{fig:standardized_data}). 
\textit{Sake} and \textit{Japanese hot pot} exhibit similar seasonality, while \textit{Japanese cuisine} shows intermittent increases, 
which may suggest a correlation with the trend of \textit{Sake}.

\begin{table*}[htbp]
\centering
\caption{Descriptive Statistics.}
\label{tab:descriptive_real}
\vspace{2mm}
\includegraphics[width=0.8\textwidth,keepaspectratio]{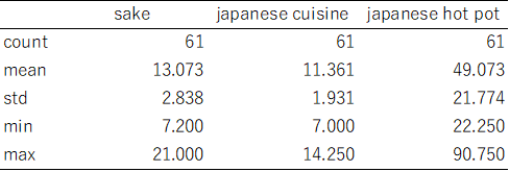}
\end{table*}

Based on these data, the models were compared (Table~\ref{tab:model_comparison_real}). 
The estimation procedure is the same as in Section~3.1, and datasets are constructed, 
split into training and test sets, and compared using the Mean Squared Error (MSE). 
The results showed that the GRU model with an \(X\) lag of 2 and a \(Z\) lag of 2 
was selected as the best-performing model.

\begin{table*}[htbp]
\centering
\caption{Model Comparison: Top 5 Results.}
\label{tab:model_comparison_real}
\vspace{2mm}
\includegraphics[width=0.5\textwidth,keepaspectratio]{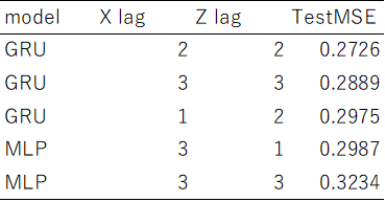}
\end{table*}

\begin{figure*}[htbp]
    \centering
    \includegraphics[width=0.5\textwidth, keepaspectratio]{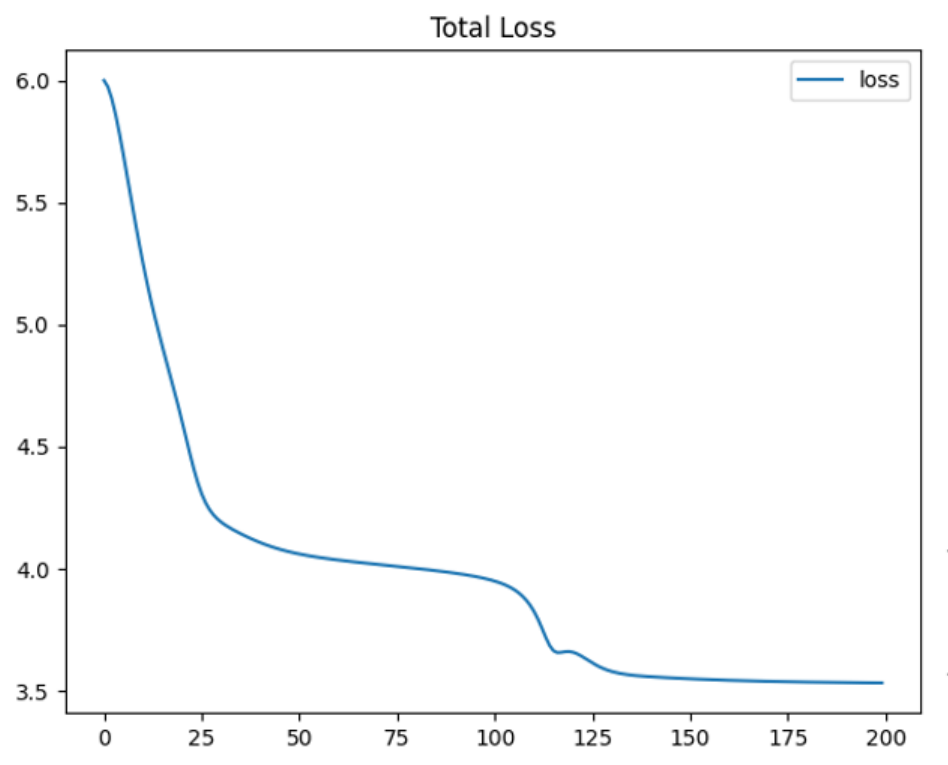}
    \caption{Convergence of optimization.}
    \label{fig:convergence}
\end{figure*}

\begin{figure*}[htbp]
    \centering
    \includegraphics[width=0.5\textwidth, keepaspectratio]{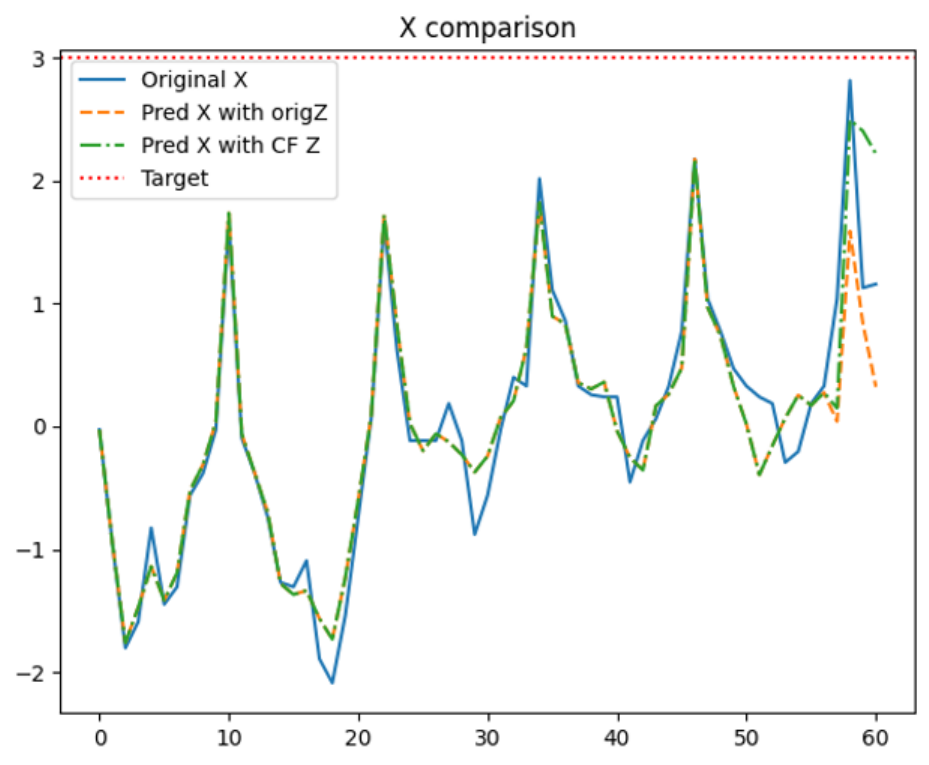}
    \caption{Actual data, model-based prediction, and prediction with counterfactual explanation (CE).}
    \label{fig:prediction_ce}
\end{figure*}

We use \(w_0\) as the weight vector \(w\), and describe the results for \(q=4\). 
First, Figure~\ref{fig:convergence} shows the optimization process for the total loss at each step. 
After a rapid decrease in the loss, the optimization temporarily plateaued before continuing to improve. 
Figure~\ref{fig:prediction_ce} shows the actual data, time series prediction, and prediction using the CE. 
It is evident that the CE-based prediction increases to match the target value even when the normal prediction typically decreases, 
demonstrating that the desired outcome is successfully achieved.

\begin{figure*}[htbp]
    \centering
    \includegraphics[width=0.9\textwidth, keepaspectratio]{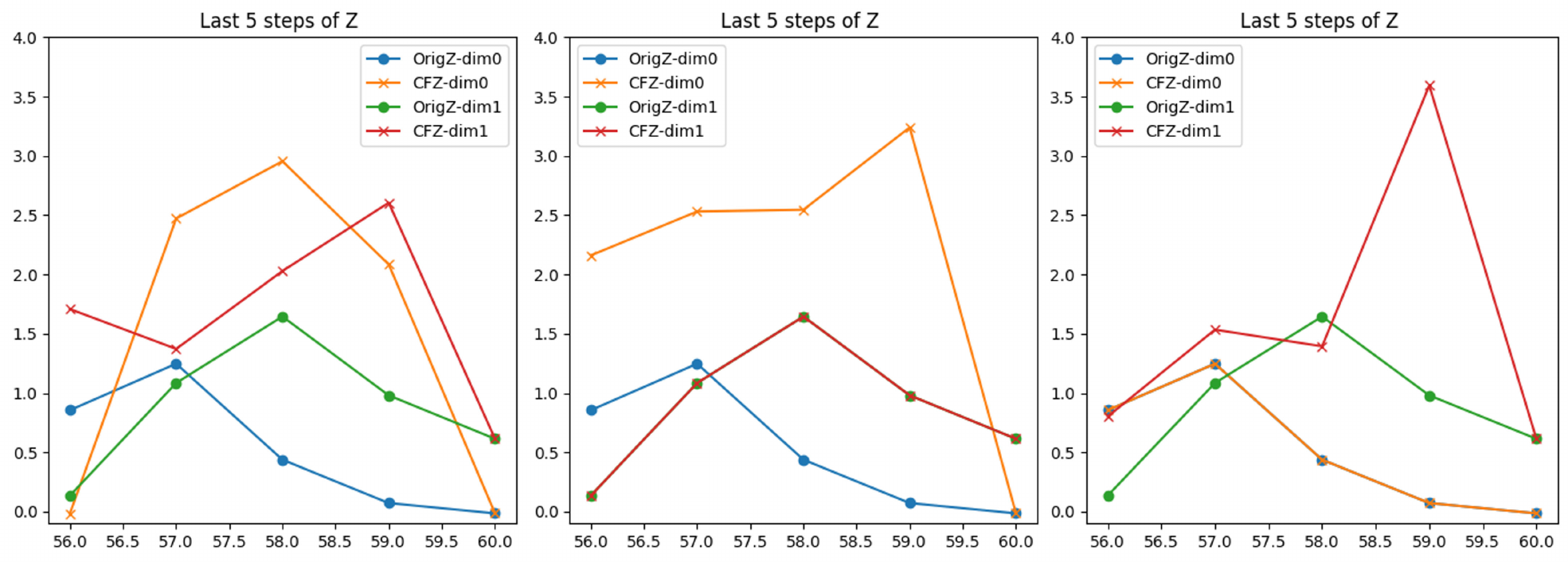}
    \caption{Time series counterfactual explanations (from left to right: both \textit{Japanese hot pot} and \textit{Japanese cuisine} optimized, only \textit{Japanese cuisine} optimized, only \textit{Japanese hot pot} optimized).}
    \label{fig:ce_timeseries}
\end{figure*}

Furthermore, we applied the method described in Section~\ref{sec2.3}, which optimizes only specific variables and compares the results. 
Figure~\ref{fig:ce_timeseries} shows three cases: 
one in which both \textit{Japanese hot pot} and \textit{Japanese cuisine} were optimized, 
one in which only \textit{Japanese cuisine} was optimized, 
and one in which only \textit{Japanese hot pot} were optimized.

\textit{Japanese cuisine} (\(k=1\)) increased steadily from the beginning. 
This suggests that its effect accumulates over time and contributes to \(x_T\) in the long term. 
For example, this can be interpreted as: “Increasing the availability of \textit{Japanese cuisine} over time ultimately boosts the sales of sake.” 
In other words, promoting \textit{Japanese cuisine} in the long term is effective for sustainably increasing sake sales. 
\textit{Japanese cuisine} (\(k=1\)) influences habits and long-term preferences.

In contrast, \textit{Japanese hot pot} (\(k=2\)) showed a noticeable increase in the most recent period. 
This suggests that it has a short-term effect and strongly contributes to the immediate value of \(x_T\). 
Since it influences \(x_T\) quickly, it can be interpreted as: 
“Increasing the availability of \textit{Japanese hot pot} in the short term immediately boosts sake sales.” 
\textit{Japanese hot pot} (\(k=2\)) may have been influenced by seasonality or a temporary mood.

When optimizing the counterfactual explanations (CE) for both \textit{Japanese cuisine} (\(k=1\)) and \textit{Japanese hot pot} (\(k=2\)), 
it is desirable to initially set \textit{Japanese cuisine} (\(k=1\)) to a low level and \textit{Japanese hot pot} (\(k=2\)) to a high level, 
and then maintain a high level of \textit{Japanese cuisine} (\(k=1\)). 
In later periods, \textit{Japanese hot pot} (\(k=2\)) was set to a lower level compared with the case where only \textit{Japanese hot pot} was optimized. 
By comparing cases where CE are extracted individually, we can better understand the interaction between \textit{Japanese cuisine} and \textit{Japanese hot pot}.

Finally, we examine the overall trends across the entire dataset (Table~\ref{tab:important_features_real}). 
On average, both \(Z_{k=1}\) and \(Z_{k=2}\) had positive effects, with the highest mean observed for \(Z_{t-2,1}\). 
However, both variables exhibit high standard deviations, indicating that their effects vary depending on the data used. 
The influence of \(Z_{t-1,2}\) and \(Z_{t-2,2}\) showed smaller standard deviations and minimum values, 
suggesting that the negative effects were relatively rare.

\begin{table*}[htbp]
\centering
\caption{Comparison of Important Features Across the Entire Time Series (Real Data).}
\label{tab:important_features_real}
\vspace{2mm}
\includegraphics[width=1.0\textwidth,keepaspectratio]{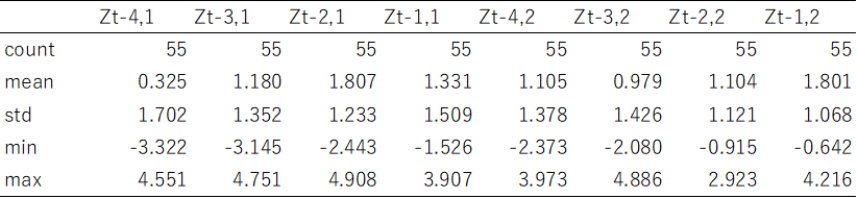}
\end{table*}

\section{Discussion and Conclusion}\label{sec4}

In this study, we propose CET-X, a method for extracting counterfactual explanations (CEs) in time series forecasting with exogenous variables. 
Additionally, we propose methods for analyzing the influence of each variable across an entire time series, detecting CEs by modifying only specific variables, 
and evaluating time series CEs. 
We also discuss the uniqueness of the solution and derive analytical solutions for specific models. 
Based on these contributions, we conducted a validation using both simulations and real-world data.

As a result, the experiments using simulated data showed that, in the case of a linear model, values close to the analytical solution could be identified. 
Furthermore, in both the linear and nonlinear models, the proposed method identified important exogenous variables. 
The effect of \(\lambda\) varied depending on \(w\), but the total loss tended to increase overall. 
Similarly, the effect of \(q\) also varied depending on \(w\); however, the total loss generally increased. 
However, in the nonlinear models, the loss in \(Z\) could sometimes decrease. 
In addition, it was revealed that better solutions could be obtained by appropriately setting the momentum in the optimization process.

These findings indicate that the combination of \(\lambda\), \(q\), and \(w\), as well as the optimization method, affects the evaluation of CE. 
Therefore, it is important to select an appropriate balance of \(\lambda\), \(q\), and \(w\), depending on the problem setting and operational goals.

Finally, in an experiment using real data, it was demonstrated that the CE can be used to bring the time series closer to the target values. 
Moreover, by comparing the CEs of individual variables and all variables, it was possible to investigate their interactions. 
Valuable insights into the strategy formulation were obtained by interpreting the outcomes.

The contributions of this study can be summarized as follows:
\begin{itemize}
    \item This study focuses on the relatively unexplored problem of counterfactual explanations (CE) in time series forecasting; 
    in particular, we propose a method to extract CE in time series forecasting with exogenous variables—common in fields such as business and marketing.
    \item It also proposes methods to analyze the influence of each variable over the entire time series, 
    extract CEs for specific variables, and evaluate time series-based CEs.
    \item Through both theoretical and data-driven validations, this study demonstrates the accuracy and practical applicability of the proposed method.
\end{itemize}

This section provides discussion and consideration on the proposed study.
Although this study assumes time point \(T\) as the final time step, 
it is also possible to generate CEs for future forecasted values \(x_{T+1},\ldots,x_{T+q}\). 
Because data for \(X\) and \(Z\) up to \(T\) are available, 
we can use them to compute the prediction \(\hat{x}_{T+1}\) and then recursively predict up to \(\hat{x}_{T+q}\).

However, when extracting CEs for future values, a challenge arises: 
there is no observed exogenous input \(Z_{T,q}\) required to calculate the distance metric in the optimization problem. 
Several strategies can be considered for addressing this issue:
\begin{enumerate}
    \item Use the most recent available value \(Z_T\),
    \item Use the average of the most recent values of \(Z\),
    \item Model the distribution or dynamics of \(Z\) (e.g., using time series forecasting) and use the predicted values for CE generation.
\end{enumerate}
These approaches provide avenues for extending \textit{CET-X} to prospective intervention scenarios.

By configuring \(w_{k,t}\), the cost of modifying \(\widetilde{Z}_{T,q}\) can be introduced in a more realistic and detailed manner. 
For instance, placing higher values on \(w_{k,t}\) at \(t=T-1\) allows us to emphasize changes in the most recent periods. 
In the context of corporate decision-making, costs that occurred further in the past may be valued more highly when converted to present value terms. 
This setting reflects these considerations.

As an extension, the proposed method can be applied to anomaly detection. 
For example, when an anomaly occurs in a time series, 
the counterfactual changes required to return the value to the normal state can be identified, thereby uncovering the cause of the anomaly.

Finally, several future challenges and directions can be outlined. 
First, from a problem-setting perspective, extending the framework to multivariate targets \(X\) is a natural next step, 
as many real-world scenarios involve multiple outcome variables. 
Second, model-related issues arise when dealing with missing or unobserved variables, 
and handling correlated error terms becomes important. 
Employing multiple models for robust estimation is another direction worth exploring. 
Lastly, improvements to the CE extraction process itself—such as using alternating optimization—could enhance the stability and interpretability of the results.
\backmatter

\bmhead{Acknowledgements}

This work was supported by JSPS KAKENHI Grant Number JP25K05381.

\begin{appendices}

\section{Proof of the Analytical Solution}\label{secA1}

We derive the optimal solution for the case where the relationship 
\(x_t = a x_{t-1} + \sum_{k=1}^{2} \beta_k z_{t-1,k} + \varepsilon_t\), 
as presented in Section~3.1, holds. 
First, we introduce the following matrices:

\(\hat{\bm{x}}\) is a \((q+1)\times1\) matrix of estimated values, 
and \(\bm{x}\) is a \((q+1)\times1\) matrix consisting of the initial values \(x_{t-q-1}\):
\[
\hat{\bm{x}}=
\begin{bmatrix}
\hat{x}_{T-q}\\
\hat{x}_{T-q+1}\\
\vdots\\
\hat{x}_T
\end{bmatrix}, 
\quad
\bm{x} = x_{T-q-1}
\begin{bmatrix}
1\\
1\\
\vdots\\
1
\end{bmatrix}.
\]

\(A\) is a \((q+1)\times(q+1)\) matrix of coefficients for \(x_t\):
\[
A =
\begin{bmatrix}
a & 0 & \cdots & 0\\
a^2 & a & \cdots & 0\\
\vdots & \vdots & \ddots & \vdots\\
a^{q+1} & a^q & \cdots & a
\end{bmatrix}.
\]

\(B = \begin{pmatrix} B_{fix} & B_{opt} \end{pmatrix}\) 
is the coefficient matrix for the intervention variable, consisting of 
the \((q+1)\times2\) matrix \(B_{fix}\) and the \((q+1)\times(2q)\) matrix \(B_{opt}\):

\[
B_{fix} =
\begin{bmatrix}
\beta_1 & \hspace{4pt} \beta_2 \\
a\beta_1 & \hspace{4pt} a\beta_2 \\
\vdots & \hspace{4pt} \vdots \\
a^{q-2}\beta_1 & \hspace{4pt} a^{q-2}\beta_2 \\
a^{q-1}\beta_1 & \hspace{4pt} a^{q-1}\beta_2
\end{bmatrix},
\quad
B_{opt} =
\begin{bmatrix}
0 & 0 & \cdots & 0 & 0 \\
\beta_1 & \hspace{4pt} \beta_2 & \cdots & 0 & 0 \\
\vdots & \hspace{4pt} \vdots & \ddots & \vdots & \vdots \\
a^{q-2}\beta_1 & \hspace{4pt} a^{q-2}\beta_2 & \cdots & 0 & 0 \\
a^{q-1}\beta_1 & \hspace{4pt} a^{q-1}\beta_2 & \cdots & \beta_1 & \hspace{4pt} \beta_2
\end{bmatrix}.
\]

\(Z = \begin{pmatrix} Z_{fix} \\ Z_{opt} \end{pmatrix}\) 
is composed of the \((2\times1)\) matrix \(Z_{fix}\) and the \((2q)\times1\) matrix \(Z_{opt}\):
\[
Z_{fix} =
\begin{bmatrix}
z_{T-q-1,1}\\
z_{T-q-1,2}
\end{bmatrix}, 
\quad
Z_{opt} =
\begin{bmatrix}
z_{T-q,1}\\
z_{T-q,2}\\
z_{T-q+1,1}\\
z_{T-q+1,2}\\
\vdots\\
z_{T-1,1}\\
z_{T-1,2}
\end{bmatrix}.
\]

Based on the above, the estimated value \(\hat{\bm{x}}\) can be computed as:
\[
\hat{\bm{x}} = A\bm{x} + B\widetilde{\bm{Z}}.
\]

If \(\bar{\bm{x}}\) is a \((q+1)\times1\) matrix with the target variables arranged vertically 
starting from \(t = T - q\), and \(W\) is a diagonal matrix with \(w_t\) as its diagonal component, 
then the optimization problem is formulated as:
\[
\widetilde{Z}_{T,q}^\ast = \widetilde{Z}_{opt}^\ast 
= \argmin_{Z_{opt}}
\Bigl\|
W^{1/2}(\bar{\bm{x}} - A\bm{x} - B_{fix}Z_{fix} - B_{opt}Z_{opt})
\Bigr\|^2
+ \lambda \| Z_{fix} - Z_{opt} \|^2.
\]

Since the fixed part \(\widetilde{Z}_{fix}\) is constant, it can be omitted.  
Let \(r = \bar{\bm{x}} - A\bm{x} - B_{fix}Z_{fix}\), and define the objective function \(J(\widetilde{Z}_{opt})\):
\[
J(\widetilde{Z}_{opt}) =
\| W^{1/2}(r - B_{opt}Z_{opt}) \|^2 + 
\lambda \| Z_{opt} - Z_{opt} \|^2.
\]

A variant of this is expressed as:
\[
J(\widetilde{Z}_{opt}) =
(r - B_{opt}\widetilde{Z}_{opt})^T W (r - B_{opt}\widetilde{Z}_{opt})
+ \lambda (\widetilde{Z}_{opt} - Z_{opt})^T(\widetilde{Z}_{opt} - Z_{opt}).
\]

By differentiating this objective function with respect to \(\widetilde{Z}_{opt}\) and setting the derivative to zero, we obtain:
\[
\frac{\partial J}{\partial \widetilde{Z}_{opt}} 
= -2B_{opt}^T W (r - B_{opt}\widetilde{Z}_{opt})
+ 2\lambda(\widetilde{Z}_{opt} - Z_{opt}) = 0.
\]

Therefore,
\[
-B_{opt}^TWr + B_{opt}^TWB_{opt}\widetilde{Z}_{opt} 
+ \lambda\widetilde{Z}_{opt} - \lambda Z_{opt} = 0.
\]

Hence,
\[
(B_{opt}^TWB_{opt} + \lambda I)\widetilde{Z}_{opt} 
= B_{opt}^TWr + \lambda Z_{opt}.
\]

From the above, the optimal value \(\widetilde{Z}_{opt}^\ast\) is given by:
\[
\widetilde{Z}_{T,q}^\ast
= \widetilde{Z}_{opt}^\ast
= (B_{opt}^TWB_{opt} + \lambda I)^{-1}
\left[
B_{opt}^TW(\bar{\bm{x}} - A\bm{x} - B_{fix}Z_{fix}) + \lambda Z_{opt}
\right].
\]

Since \(J(\widetilde{Z}_{opt})\) is formulated as a quadratic function consisting of a squared error term and a ridge regularization term, the objective function becomes convex if \(W\) is positive definite and \(\lambda > 0\), ensuring that this solution yields the global minimum. \(\square\)

Moreover, this result holds when generalized to higher-order lags \(m\) for \(x_{t-m}\) and \(n\) for \(z_{t-n,k}\) as long as the model remains linear.

\end{appendices}


\bibliography{sn-bibliography}% common bib file

@article{angelov2021explainable,
  author = {Angelov, P. P. and Soares, E. A. and Jiang, R. and Arnold, N. I. and Atkinson, P. M.},
  title = {Explainable artificial intelligence: an analytical review},
  journal = {Wiley Interdisciplinary Reviews: Data Mining and Knowledge Discovery},
  year = {2021},
  volume = {11},
  number = {5},
  pages = {e1424}
}

@article{minh2022explainable,
  author = {Minh, D. and Wang, H. X. and Li, Y. F. and Nguyen, T. N.},
  title = {Explainable artificial intelligence: a comprehensive review},
  journal = {Artificial Intelligence Review},
  year = {2022},
  pages = {1--66}
}

@article{dwivedi2023explainable,
  author = {Dwivedi, R. and Dave, D. and Naik, H. and Singhal, S. and Omer, R. and Patel, P. and Ranjan, R.},
  title = {Explainable AI (XAI): Core ideas, techniques, and solutions},
  journal = {ACM Computing Surveys},
  year = {2023},
  volume = {55},
  number = {9},
  pages = {1--33}
}

@article{saeed2023explainable,
  author = {Saeed, W. and Omlin, C.},
  title = {Explainable AI (XAI): A systematic meta-survey of current challenges and future opportunities},
  journal = {Knowledge-Based Systems},
  year = {2023},
  volume = {263},
  pages = {110273}
}

@article{du2019techniques,
  author = {Du, M. and Liu, N. and Hu, X.},
  title = {Techniques for interpretable machine learning},
  journal = {Communications of the ACM},
  year = {2019},
  volume = {63},
  number = {1},
  pages = {68--77}
}

@book{molnar2020interpretable,
  author = {Molnar, C.},
  title = {Interpretable machine learning},
  year = {2020},
  publisher = {Lulu.com},
  address   = {United States}
}

@article{theissler2022explainable,
  author = {Theissler, A. and Spinnato, F. and Schlegel, U. and Guidotti, R.},
  title = {Explainable AI for time series classification: a review, taxonomy and research directions},
  journal = {IEEE Access},
  year = {2022},
  volume = {10},
  pages = {100700--100724}
}

@inproceedings{bento2021timeshap,
  author = {Bento, J. and Saleiro, P. and Cruz, A. F. and Figueiredo, M. A. and Bizarro, P.},
  title = {Timeshap: Explaining recurrent models through sequence perturbations},
  booktitle = {Proceedings of the 27th ACM SIGKDD Conference on Knowledge Discovery \& Data Mining},
  year = {2021},
  pages = {2565--2573}
}

@article{spinnato2023understanding,
  author = {Spinnato, F. and Guidotti, R. and Monreale, A. and Nanni, M. and Pedreschi, D. and Giannotti, F.},
  title = {Understanding any time series classifier with a subsequence-based explainer},
  journal = {ACM Transactions on Knowledge Discovery from Data},
  year = {2023},
  volume = {18},
  number = {2},
  pages = {1--34}
}

@article{verma2024counterfactual,
  author = {Verma, S. and Boonsanong, V. and Hoang, M. and Hines, K. and Dickerson, J. and Shah, C.},
  title = {Counterfactual explanations and algorithmic recourses for machine learning: A review},
  journal = {ACM Computing Surveys},
  year = {2024},
  volume = {56},
  number = {12},
  pages = {1--42}
}

@article{guidotti2024counterfactual,
  author = {Guidotti, R.},
  title = {Counterfactual explanations and how to find them: literature review and benchmarking},
  journal = {Data Mining and Knowledge Discovery},
  year = {2024},
  volume = {38},
  number = {5},
  pages = {2770--2824}
}

@article{wachter2017counterfactual,
  author = {Wachter, S. and Mittelstadt, B. and Russell, C.},
  title = {Counterfactual explanations without opening the black box: Automated decisions and the GDPR},
  journal = {Harvard Journal of Law and Technology},
  year = {2017},
  volume = {31},
  pages = {841}
}

@inproceedings{goyal2019counterfactual,
  author = {Goyal, Y. and Wu, Z. and Ernst, J. and Batra, D. and Parikh, D. and Lee, S.},
  title = {Counterfactual visual explanations},
  booktitle = {International Conference on Machine Learning},
  year = {2019},
  pages = {2376--2384},
  organization = {PMLR}
}

@inproceedings{ates2021counterfactual,
  author = {Ates, E. and Aksar, B. and Leung, V. J. and Coskun, A. K.},
  title = {Counterfactual explanations for multivariate time series},
  booktitle = {2021 International Conference on Applied Artificial Intelligence (ICAPAI)},
  year = {2021},
  pages = {1--8},
  organization = {IEEE}
}

@article{prado2024survey,
  author = {Prado-Romero, M. A. and Prenkaj, B. and Stilo, G. and Giannotti, F.},
  title = {A survey on graph counterfactual explanations: definitions, methods, evaluation, and research challenges},
  journal = {ACM Computing Surveys},
  year = {2024},
  volume = {56},
  number = {7},
  pages = {1--37}
}

@inproceedings{jeanneret2024text,
  author = {Jeanneret, G. and Simon, L. and Jurie, F.},
  title = {Text-to-image models for counterfactual explanations: a black-box approach},
  booktitle = {Proceedings of the IEEE/CVF Winter Conference on Applications of Computer Vision},
  year = {2024},
  pages = {4757--4767}
}

@inproceedings{delaney2021instance,
  author = {Delaney, E. and Greene, D. and Keane, M. T.},
  title = {Instance-based counterfactual explanations for time series classification},
  booktitle = {International Conference on Case-Based Reasoning},
  year = {2021},
  pages = {32--47},
  publisher = {Springer International Publishing},
  address = {Cham}
}

@inproceedings{li2022sgcf,
  author = {Li, P. and Bahri, O. and Boubrahimi, S. F. and Hamdi, S. M.},
  title = {SG-CF: Shapelet-guided counterfactual explanation for time series classification},
  booktitle = {2022 IEEE International Conference on Big Data (Big Data)},
  year = {2022},
  pages = {1564--1569},
  organization = {IEEE}
}

@article{bahri2022shapelet,
  author = {Bahri, O. and Boubrahimi, S. F. and Hamdi, S. M.},
  title = {Shapelet-based counterfactual explanations for multivariate time series},
  journal = {arXiv preprint arXiv:2208.10462},
  year = {2022}
}

@inproceedings{refoyo2024subspace,
  author = {Refoyo, M. and Luengo, D.},
  title = {Sub-SpaCE: Subsequence-Based Sparse Counterfactual Explanations for Time Series Classification Problems},
  booktitle = {World Conference on Explainable Artificial Intelligence},
  year = {2024},
  pages = {3--17},
  publisher = {Springer Nature Switzerland},
  address = {Cham}
}

@inproceedings{bahri2025denoising,
  author = {Bahri, O. and Li, P. and Hosseinzadeh, P. and Boubrahimi, S. F. and Hamdi, S. M.},
  title = {Denoising Optimization-Based Counterfactual Explanations for Time Series Classification},
  booktitle = {International Conference on Pattern Recognition},
  year = {2025},
  pages = {162--179},
  publisher = {Springer},
  address = {Cham}
}

@inproceedings{filali2022barycenter,
  author = {Filali Boubrahimi, S. and Hamdi, S. M.},
  title = {On the mining of time series data counterfactual explanations using barycenters},
  booktitle = {Proceedings of the 31st ACM International Conference on Information \& Knowledge Management},
  year = {2022},
  pages = {3943--3947}
}

@inproceedings{hollig2022tsevo,
  author = {Höllig, J. and Kulbach, C. and Thoma, S.},
  title = {Tsevo: Evolutionary counterfactual explanations for time series classification},
  booktitle = {2022 21st IEEE International Conference on Machine Learning and Applications (ICMLA)},
  year = {2022},
  pages = {29--36},
  organization = {IEEE}
}

@article{huang2024txgen,
  author = {Huang, Q. and Kitharidis, S. and Bäck, T. and van Stein, N.},
  title = {TX-Gen: Multi-Objective Optimization for Sparse Counterfactual Explanations for Time-Series Classification},
  journal = {arXiv preprint arXiv:2409.09461},
  year = {2024}
}

@article{refoyo2024multispace,
  author = {Refoyo, M. and Luengo, D.},
  title = {Multi-SpaCE: Multi-Objective Subsequence-based Sparse Counterfactual Explanations for Multivariate Time Series Classification},
  journal = {arXiv preprint arXiv:2501.04009},
  year = {2024}
}

@article{wang2024glacier,
  author = {Wang, Z. and Samsten, I. and Miliou, I. and Mochaourab, R. and Papapetrou, P.},
  title = {Glacier: guided locally constrained counterfactual explanations for time series classification},
  journal = {Machine Learning},
  year = {2024},
  volume = {113},
  number = {7},
  pages = {4639--4669}
}

@inproceedings{li2023cels,
  author = {Li, P. and Bahri, O. and Boubrahimi, S. F. and Hamdi, S. M.},
  title = {Cels: Counterfactual explanations for time series data via learned saliency maps},
  booktitle = {2023 IEEE International Conference on Big Data (BigData)},
  year = {2023},
  pages = {718--727},
  organization = {IEEE}
}

@article{li2024mcels,
  author = {Li, P. and Bahri, O. and Boubrahimi, S. F. and Hamdi, S. M.},
  title = {M-CELS: Counterfactual Explanation for Multivariate Time Series Data Guided by Learned Saliency Maps},
  journal = {arXiv preprint arXiv:2411.02649},
  year = {2024}
}

@inproceedings{yan2023selfinterpretable,
  author = {Yan, J. and Wang, H.},
  title = {Self-interpretable time series prediction with counterfactual explanations},
  booktitle = {International Conference on Machine Learning},
  year = {2023},
  pages = {39110--39125},
  organization = {PMLR}
}

@article{sun2024counterfactual,
  author = {Sun, X. and Aoki, R. and Wilson, K. H.},
  title = {Counterfactual Explanations for Multivariate Time-Series without Training Datasets},
  journal = {arXiv preprint arXiv:2405.18563},
  year = {2024}
}

@inproceedings{wang2023forecasting,
  author = {Wang, Z. and Miliou, I. and Samsten, I. and Papapetrou, P.},
  title = {Counterfactual explanations for time series forecasting},
  booktitle = {2023 IEEE International Conference on Data Mining (ICDM)},
  year = {2023},
  pages = {1391--1396},
  organization = {IEEE}
}

@article{sulem2022diverse,
  author = {Sulem, D. and Donini, M. and Zafar, M. B. and Aubet, F. X. and Gasthaus, J. and Januschowski, T. and Archambeau, C.},
  title = {Diverse counterfactual explanations for anomaly detection in time series},
  journal = {arXiv preprint arXiv:2203.11103},
  year = {2022}
}

@inproceedings{tsirtsis2021counterfactual,
  author = {Tsirtsis, S. and De, A. and Rodriguez, M.},
  title = {Counterfactual explanations in sequential decision making under uncertainty},
  booktitle = {Advances in Neural Information Processing Systems},
  year = {2021},
  volume = {34},
  pages = {30127--30139}
}

@article{hochreiter1997long,
  author = {Hochreiter, S. and Schmidhuber, J.},
  title = {Long short-term memory},
  journal = {Neural Computation},
  year = {1997},
  volume = {9},
  number = {8},
  pages = {1735--1780}
}

@article{chung2014empirical,
  author = {Chung, J. and Gulcehre, C. and Cho, K. and Bengio, Y.},
  title = {Empirical evaluation of gated recurrent neural networks on sequence modeling},
  journal = {arXiv preprint arXiv:1412.3555},
  year = {2014}
}

\end{document}